\documentclass{article}

\PassOptionsToPackage{numbers, compress}{natbib}

\usepackage[final]{neurips_2019}

\usepackage[utf8]{inputenc} 
\usepackage[T1]{fontenc}    
\usepackage{hyperref}       
\usepackage{url}            
\usepackage{booktabs}       
\usepackage{amsfonts}       
\usepackage{nicefrac}       
\usepackage{microtype}      
\usepackage{amsmath}
\usepackage{tikz}
\usepackage{color}
\usepackage{amssymb}
\usepackage{thmtools}
\usepackage{thm-restate}

\usetikzlibrary{bayesnet}
\usepackage[font=small,labelfont=bf]{caption}
\usepackage{subcaption}
\usepackage{amsthm}

\theoremstyle{definition}
\newtheorem{definition}{Definition}[section]

\theoremstyle{lemma}

\newtheorem{corollary}{Corollary}

\newcommand{\R}{\mathbb{R}}
\newcommand{\E}{\mathbb{E}}
\newcommand{\bX}{\mathbf{X}}
\newcommand{\bx}{\mathbf{x}}
\newcommand{\bU}{\mathbf{U}}
\newcommand{\bu}{\mathbf{u}}
\newcommand{\bS}{\mathbf{S}}
\newcommand{\bs}{\mathbf{s}}
\newcommand{\bZ}{\mathbf{Z}}

\newcommand{\bmu}{\pmb{\mu}}
\newcommand{\calX}{\mathcal{X}}

\newcommand{\calS}{\mathcal{S}}
\newcommand{\calD}{\mathcal{D}}
\newcommand{\calZ}{\mathcal{Z}}
\newcommand{\calN}{\mathcal{N}}
\newcommand{\calU}{\mathcal{U}}

\title{
Location Trace Privacy Under Conditional Priors}

\author{%
  Casey Meehan \\
  University of California, San Diego\\
  \texttt{cmeehan@eng.ucsd.edu} \\
   \And
   Kamalika Chaudhuri \\
   University of California, San Diego \\
   \texttt{kamalika@eng.ucsd.edu} \\
}

\begin{document}

\maketitle

\begin{abstract}
    Providing meaningful privacy to users of location based services is particularly challenging when multiple locations are revealed in a short period of time. This is primarily due to the tremendous degree of dependence that can be anticipated between points. We propose a R\'enyi differentially private framework for bounding expected privacy loss for conditionally dependent data. Additionally, we demonstrate an algorithm for achieving this privacy under Gaussian process conditional priors. This framework both exemplifies why conditionally dependent data is so challenging to protect and offers a strategy for preserving privacy to within a fixed radius for every user location in a trace. 
\end{abstract}
\section{Introduction}
Location data is acutely sensitive information, detailing where we live, work, eat, shop, worship, and often when, too. Yet increasingly, location data is being uploaded for smartphone services such as ride hailing and weather forecasting and then being brokered in a thriving user location aftermarket to advertisers and even investors \cite{nyt}. As such, there is a keen need for privacy preserving methods with meaningful guarantees. However, for virtually any application involving locations that are \emph{dependent}, like location traces, there are few results in the privacy literature. 

 The widely accepted standard of \emph{Geo-Indistinguishability} (GI) \citep{GI} effectively applies differential privacy to location points and makes use of the Laplace mechanism for sanitizing them. The idea is that a given point in space enjoys a radius of privacy after noise is added, wherein the radius and degree of privacy are tuneable parameters. However, like most privacy literature GI is prior agnostic, meaning it ignores any prior dependence between points in a user's location trace. Even under weak conditional dependence, the privacy guarantees begin to quickly dissolve. 
 
A number of existing works have considered the problem setting of aggregate location data and queries. Here, individual data is protected in a dataset of aggregated location trajectories over a large number of people and an adversary making aggregate data queries about the dataset \cite{traffic_monitoring, aggregated_encryption}. These, however, provide little utility when applied to a single or a few location traces. Another, more general, line of work such as the median and matrix mechanisms aim to provide privacy when a series of correlated and aggregate \emph{queries} are made on the same database, allowing us to provide prior-agnostic solutions which offer both privacy and utility \cite{roth, matrix_mechanism}. These too, focus on aggregate data and consider protecting against correlated queries, as opposed to correlated data points.

Other work perhaps closer to our setting is that of inferential privacy such as Pufferfish privacy \cite{pufferfish}. More general works focused on inferential privacy such as \cite{markov, universal} fail to address the spatiotemporal aspects of the problem. Inferential privacy methods more specifically aimed at location trace privacy such as \cite{temporal, predictive}, are either too limiting in their prior distribution, or only consider on-line location data release as opposed to full trace release. 

In this work, we revisit the problem of preserving privacy for a single user's location trace under the assumption of a flexible conditional prior. We first lay out a Pufferfish style framework that limits the class of prior distributions and the discriminative pairs to be protected. We then specify a R\'enyi differential privacy measurement \citep{renyi} of privacy loss, and finally show how to bound this loss under Gaussian assumptions, thus preserving privacy to within a specified radius of each point in the trace. Finally, we provide some empirical examples of how even under weak correlative priors, more noise is needed than would be specified by prior agnostic approaches such as GI. 

\section{Preliminaries and Problem Formulation}

\begin{figure*}[h]
	\centering
	\begin{subfigure}{.45\textwidth}
		\centering
     \scalebox{0.8}{\begin{tikzpicture}

  \node[obs]                       (x1) {$x_1$};
  \node[latent, below=of x1]			  (z1) {$z_1$};
  \node[obs, right=of x1]             (x2) {$x_2$};
  \node[latent, below=of x2]			  (z2) {$z_2$};
  \node[latent, right=of x2]          (x3) {$x_3$};
  \node[latent, below=of x3]			  (z3) {$z_3$};
  \node[latent, right=of x3]          (x4) {$x_4$};
  \node[latent, below=of x4]			  (z4) {$z_4$};

  \edge {x1} {z1};
  \edge {x2} {z2}; 
  \edge {x3} {z3};
  \edge {x4} {z4};
  \edge [-] {x1} {x2}; 
  \edge [-] {x2} {x3};
  \edge [-] {x3} {x4};
  \draw (x1) edge[-, bend left = 40] node [right] {} (x3);
  \draw (x1) edge[-, bend left = 40] node [right] {} (x4);
  \draw (x2) edge[-, bend left = 40] node [right] {} (x4);


\end{tikzpicture}}  

		\caption{}
		\label{full model}
	\end{subfigure}
	\begin{subfigure}{.45\textwidth}
		\centering
     \scalebox{0.8}{\begin{tikzpicture}

  \node[obs]                       (s) {$\mathbf{S},\mathcal{S}$};
  \node[latent, below=of x1]			  (zs) {$\mathbf{Z}_s, \mathcal{Z}_s$};
  \node[latent, right=of x1]             (u) {$\mathbf{U}, \mathcal{U}$};
  \node[latent, below=of x2]			  (zu) {$\mathbf{Z}_u, \mathcal{Z}_u$};

  \edge {s} {zs};
  \edge {u} {zu}; 
  \edge [-] {s} {u}; 


\end{tikzpicture}}  

		\caption{}
		\label{condensed model}
	\end{subfigure}
	\caption{(a) An example graphical model of a four point trace $\bX \in \R^4$. (b) The more general grouped version of the model in (a), where the secret set $\calS = \{x_1, x_3\}$, the remaining set $\calU = \{x_3, x_4\}$, and the obfuscated $\calZ_u$, $\calZ_s$ are shown.}
	\label{graphical models}
\end{figure*}
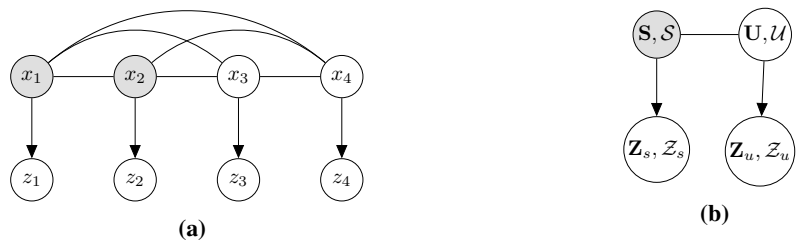

Without loss of generality, we consider a location trace as $d$ real-valued points. We refer to a $d$-point location trace as either a vector in $\R^d$, $\bX = [x_1, \dots, x_d]$, or a set of $d$ points in $\R$, $\calX = \{x_1, \dots, x_d\}$. We will use these two definitions interchangeably depending on the context. Similarly, we refer to the $d$-point privatized version of $\bX, \calX$ as $\bZ, \calZ$, existing in the same instance space. A mechanism $M$ privatizes each $x_i \in \calX$ to its corresponding $z_i \in \calZ$ as i s depicted in the graphical model of \textbf{ Figure \ref{full model} }. For a more realistic 2- or 3-dimensional application, either the dimensions may be deemed independent and privatized separately, or be linked through techniques like co-krigging. 

In the same fashion as GI, we aim to provide privacy to within a radius $r$ for each real-valued location point $x_i \in \calX$. However, existing location privacy notions like GI consider a mechanism $M(\bX) = \bZ$ sufficient if each $z_i \in \calZ$ does not relate where its corresponding $x_i \in \calX$ is within some radius $r$. In this work we motivate the need for a more strict notion of location privacy that requires each $z_i$ to not only protect its corresponding $x_i$, but what might be implied about the remaining $\calX \backslash x_i$. Additionally, $M$ must privatize subsequences of points $\calS \in \calX$ by ensuring the released subsequence $\calZ_s \in \calZ$ does not relate where $\calS$ \emph{or} the remaining (complement) subsequence points $\calU = \calX \backslash \calS$ are within radius $r$. 

To formalize what we mean by $\calS$ `implying' something about $\calU$, we mean that it significantly affects the conditional probability $P(\bU| \bS)$. We ask that any privacy preserving mechanism be private to some class of `conditional priors'. 
\begin{definition}
	A class of conditional priors specifies a set, $\calD$, of conditional prior distributions, $\theta \in \calD$, that define the probability of any subsequence of location points $\calU \in \calX$ conditioned on the remaining points $\calS = \calX \backslash \calU$: $P(\bU = \bu | \bS = \bs, \theta)$. 
\end{definition}
It is important to note that specifying the class of conditional priors still leaves a large class of marginal priors $P(\bX = \bx | \theta)$. For example, a Gaussian conditional prior does not require a Gaussian marginal prior. 

To see the importance of subsequence privacy, consider a 3-point trace, $\calX_3 = \{x_1, x_2, x_3\}$, and its privatized trace $\calZ$. We reason about the odds of the points being in the locations $x_{\text{daycare}}$, $x_{\text{work}}$ which are within $r$ of each other. Frameworks like GI assume independence of each $x_i \in \calX$, and then bound the odds of any $z_i$ given $x_i$ is at $x_{\text{daycare}}$ versus $x_{\text{work}}$. However, if the $x_i$'s are interdependent (and likely are) this may not be the case. Reasoning about $\calS = \{x_1, x_3\}$ and $\calU = \calX \backslash \calS = \{x_2\}$, a mechanism satisfying GI does not bound the odds of $\calZ$ given $x_1 = x_{\text{daycare}} \cap x_3 = x_{\text{work}}$ versus $x_1 = x_{\text{work}} \cap x_3 = x_{\text{daycare}}$. Depending on the conditional prior $\theta \in \calD$, these two orders of events may have very different implications about the location of $x_2$, and thus $z_2$. 

There are two primary distinguishing features about our location privacy framework. First, instead of individually bounding the odds of $\frac{P(z_i | x_i = x')}{P(z_i | x_i = x)}$ for pairs of $x',x$ hypotheses within $r$, we bound the odds of $\frac{P(\bZ | \bS = \bs')}{P(\bZ | \bS = \bs)}$ for pairs of $\bs', \bs$ hypotheses such that each hypothesis pair $x_i', x_i \in \calS$ are within $r$. Second, instead of bounding max divergence as in GI, we opt to bound the $\lambda$-th order R\'enyi divergence for simplicity of analysis with a wide range of conditional prior classes, $\calD$. Recall that, for distributions $\mu, \nu$ on the same measurable space, this is defined as $D_\lambda \binom{\mu}{\nu} = \frac{1}{\lambda - 1} \log \E_{x \sim \nu} \bigg( \frac{\mu(x)}{\nu(x)} \bigg)^\lambda$. 

Borrowing from `Pufferfish Privacy' defined in \cite{pufferfish}, we guarantee privacy for a specific set of secret pairs of subsequence value assignments $(\bS = \bs_i, \bS = \bs_j)$, and for a specific (conditional) prior class $\calD$: 

\begin{definition}{ $\big[$ ($\epsilon, r, \calD$)-Conditional Inferential Privacy (CIP) $\big]$}
	A randomized mechanism $M:\R^d \rightarrow \R^d$ applied to location trace $\bX, \calX$ offers subsequence $\calS \subseteq \calX$ ($\epsilon, r, \calD$)-CIP provided that 
	\begin{itemize}
		\item for all possible outputs $\bZ \in \R^d$
		\item for all discriminative pairs $\calS_{\text{pairs}} = \{ (\bs_i, \bs_j) : \| \bs_i - \bs_j \|_\infty \leq r\}$ for subsequence $\calS \subseteq \calX$
		\item for all conditional priors $\theta \in \calD$ such that $P(\bS = \bs_i | \theta) \neq 0$, $P(\bS = \bs_j | \theta) \neq 0$ for all discriminative pairs 
	\end{itemize}	
	\begin{align*}
		D_{\lambda} \binom{P(\bZ | \bS = \bs_i, \theta)}{P(\bZ | \bS = \bs_j, \theta)} \leq \epsilon
	\end{align*}
	\label{CIP def}
\end{definition} 


Where for simplicity we will refer to the events $\bS = \bs_i$, $\bS = \bs_j$, as $\bS_i$, $\bS_j$, respectively.

An interpretation of ($\epsilon, r, \calD$)-CIP compliance for subsequence $\calS$ is that release $\calZ$ does not sharpen any marginal prior (that has a conditional prior in $\calD$) by more than $\epsilon$ in expectation. 

\begin{corollary}
	\label{expected_gap}
	For two distributions, $P(\bZ | \bS_i, \theta), P(\bZ | \bS_j, \theta)$, with R\'enyi Divergence bounded by $\epsilon$, we also know that the expected gap between the prior and posterior log-odds ratios is bounded by $\epsilon$. 
	\begin{align}
		\E_{\bZ \sim P(\bZ | \bS_j, \theta)} \bigg[ \log \frac{P(\bS_i | \bZ, \theta)}{P(\bS_j | \bZ, \theta)} - \log \frac{P(\bS_i | \theta)}{P(\bS_j | \theta)} \bigg] 
		&\leq D_{\lambda} \binom{P(\bZ | \bS_i, \theta)}{P(\bZ | \bS_j, \theta)} \label{odds bound} \\
		&\leq \epsilon \notag
	\end{align}
\end{corollary}

Returning to the original intention of providing privacy within radius $r$ of \emph{each} $x_i \in \calX$, we point out that a CIP-compliant mechanism, $M$, for subsequence $\calS$ does not necessarily guarantee this radius to each $x_i \in \calS$. This is because --- without any constraint on $\calD$ --- there may exist another subsequence including $x_i$, $\calS' \supseteq \{x_i\}$, for which $M$ is not CIP compliant. As such, we assert that achieving true location privacy for point $x_i$ in trace $\calX$, requires being CIP compliant for every subsequence containing $x_i$, which is exponential in the number of points $|\calX|$. This exemplifies why privacy for highly dependent data is challenging: the number of ways an adversary can infer the true value of $x_i$ is exponential in the number of points. 

In the next section, we demonstrate a CIP compliant mechanism for a Gaussian process conditional prior and additive Gaussian mechanism.

\section{CIP Compliance Under Gaussian Assumptions}

\subsection{The Privacy Bound for Additive Mechanisms}
It is instructive to analyze the privacy loss in the case of an additive noise mechanism, $\bZ \sim M(\bX) = g + \bX$, where $g \sim \zeta$. Using the model depicted in Figure \ref{graphical models}, we know that all of the $x \in \calX$ values are fully connected, and each of the $z_i \in \calZ$ values are conditionally independent given their associated $x_i$ value. Using this conditional independence relation, reconsider the R\'enyi Divergence bound defined in Definition \ref{CIP def}. Here, the notation $g_s, g_u$ correspond to the noise vector $g$ indexed at the $\bS$ and $\bU$ points, respectively. 
\begin{align} 
	\label{div terms} 
	D_\lambda \binom{P(\bZ | \bS_i, \theta)}{P(\bZ | \bS_j, \theta)}  
	&= D_\lambda \binom{P(\bU_i + g_u)}{P(\bU_j + g_u)}  +
	\sum_{(z, x_i, x_j) \in (\calZ_s, \calS_i, \calS_j)} 
	D_\lambda \binom{P(z | x_i)}{P(z | x_j)}  
\end{align}
where $\bU_i \sim P(\bU | \bS_i, \theta), \bU_j \sim P(\bU | \bS_j, \theta)$. (See Appendix for proof)

Here, we see that the divergence separates into two terms: the first for $\calZ_u$ and the second for $\calZ_s$. The second is prior agnostic, behaving much more like differential privacy. The first term depends on the conditional distribution, $\theta$, of the $\bU$ points on the $\bS$ points and on the mechanism noise $g \sim \zeta$.

\subsection{CIP for Gaussian Process Data}
We now demonstrate a CIP mechanism for the case of a Gaussian process conditional prior class, $\theta \in \calD$, where a timestamp $t_i$ is released for each $x_i \in \calX$. The conditional distribution for any partition of $\calX$, $P(\bU | \bS)$ is the conditional distribution of a mean zero multivariate normal distribution with some covariance matrix $\Sigma$ dictated by some kernel $\Sigma_{ij} = k(t_i, t_j)$. 
\begin{align}
	P(\bU | \bS) &= \calN(\bmu_{u|s}, \Sigma_{u|s}) \notag 
\end{align}
where $\bmu_{u|s} = \Sigma_{us} \Sigma_{ss}^{-1} \bS$ and $\Sigma_{u|s} = \Sigma_{uu} - \Sigma_{us}\Sigma_{ss}^{-1}\Sigma_{su}$. The class of covariance matrices, $\calD$, includes all covariance matrices of an RBF kernel with equal variance $\sigma_x^2$ and with length scales less than some maximum length scale $l \leq l_{\text{max}}$. For location points with time values $t_1$ and $t_2$, $\Sigma_{ij} = k_{\text{RBF}}(t_i, t_j) = \sigma_x^2 \exp \bigg( \frac{-(t_i - t_j)^2}{2 l^2} \bigg)$. However, this analysis applies to any kernel. As a first pass, we consider the mechanism of adding mean zero, equal variance, Gaussian noise to each location point: $\bZ = \bX + g$ and $g \sim \calN(0, \sigma_z^2 \mathbb{I})$. 

While $\bX$ is not necessarily Gaussian distributed, $P(\bZ | \bS_i, \theta) = P(\bZ_s, \bZ_u | \bS_i, \theta) = P(\bZ_s | \bS_i)P(\bU_i + g_u)$ \emph{is} Gaussian distributed, making the R\'enyi Divegence in Definition \ref{CIP def} easy to analyze. Substituting this fact into Equation \ref{div terms}, we solve four our privacy loss, $\mathbf{L}_{\text{priv}}$ 
\begin{align}
	\mathbf{L}_{\text{priv}}
	=& \frac{\lambda}{2} \Big[ \Delta \bs^\intercal \Sigma_{\textit{eff}} \Delta \bs + 
	\frac{1}{\sigma_z^2} \| \Delta \bs \|_2^2  \Big] \label{2terms}
\end{align}
where $\Delta \bs = (\bs_i - \bs_j)$ and $\Sigma_{\textit{eff}} = (\Sigma_{us}\Sigma_{ss}^{-1})^\intercal (\Sigma_{u|s} + \sigma_z \mathbb{I})^{-1} (\Sigma_{us}\Sigma_{ss}^{-1})$. (See Appendix for proof)

By Definition \ref{CIP def}, $\calS_{\text{pairs}}$ requires that $\| \Delta \bs \|_\infty \leq r$. For simplicity, we show maximum privacy loss for a wider set discriminative pairs: $\calS_{\text{pairs}} = \{\Delta \bs \in \R^{|\calS|} : \| \Delta \bs \|_2^2 \leq  r^2 |\calS|\}$, effectively solving for the worst case privacy loss in the $L_2$ ball circumscribing the $L_\infty$ ball suggested in Definition \ref{CIP def}. 

This allows us to easily maximize the fist term in Equation \ref{2terms}: a Mahalanobis distance maximized by the eigenvector of $\Sigma_\textit{eff}$ with maximal eigenvalue. The second term, being prior agnostic and only affected by magnitude of $\Delta \bs$, will automatically be maximized if the first term is maximized. We denote this worst case eigenvector discriminative pair $\Delta \bs^*$, and associated worst case loss as $\mathbf{L}_{\text{priv}}^*$, which may be simplified as
\begin{align}
	\mathbf{L}_{\text{priv}}^* &= \frac{\lambda}{2} \bigg( \frac{1 + \sigma_z^2 \alpha^*}{\sigma_z^2} \bigg) |\calS|  r^2 \label{Lpriv_star}
\end{align}
where $\alpha^*$ is the largest eigenvalue of $\Sigma_{\textit{eff}}$. If we choose a $\sigma_z^2$ large enough that $\mathbf{L}_{\text{priv}}^* \leq \epsilon$, then the mechanism is CIP compliant for subsequence $\calS$. Furthermore, if the mechanism is CIP compliant for the covariance matrix with highest correlation induced by $l = l_{\text{max}}$, then it is CIP compliant for all $l < l_{\text{max}}$, and thus all $\theta \in \calD$. 

\begin{figure*}[h]
	\centering
	\begin{subfigure}[b]{0.45 \textwidth}
		\includegraphics[width = \textwidth]{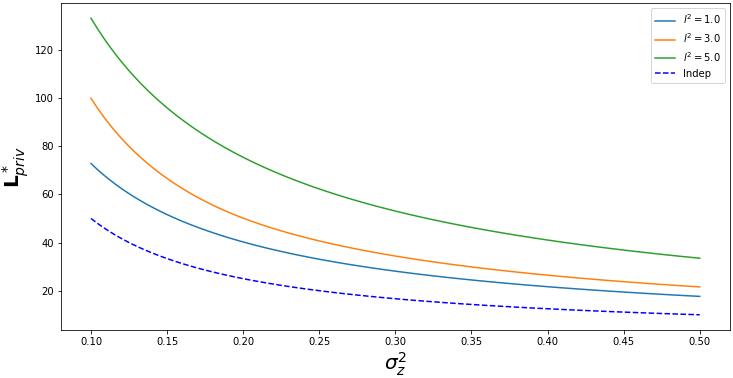}
		\label{scale sigma}
		\caption{}
	\end{subfigure}
	\hfill
	\begin{subfigure}[b]{0.45 \textwidth}
		\includegraphics[width = \textwidth]{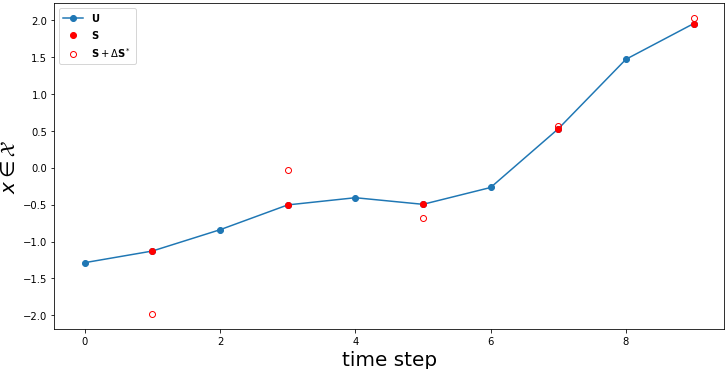}
		\label{example trace}
		\caption{}
	\end{subfigure}
	\caption{\textbf{(a)} Worst case privacy loss vs. noise variance, $\sigma_z^2$, added to each $x_i$ for a subsequence $\calS$ of every other point in $\calX$. We show privacy loss of several RBF kernel covariance matrices $\Sigma$ with different length scales $l^2$. The blue dotted line shows the privacy loss for points that are assumed independent $\Sigma = \mathbb{I}$ (as in GI), which drastically under-represents privacy loss against an RBF conditional prior. \textbf{(b)} Here, we show an example location trace, $\calX$, and superimpose the worst case hypothesis on $\Delta \bs^*$ for an RBF conditional prior $\theta$ of length scale $l = 2$. If an adversary compares $\bS = \bs_i$ with $\bS = \bs_j = \bs_i + \Delta \bs^*$, they will witness a large deviation in the distribution of the remaining points $\bZ_u$.}
	\label{gaussian figs}
\end{figure*}

Prior agnostic methods like GI, effectively assume all points are independent, and thus $\Sigma = \mathbb{I}$. To demonstrate, we compute the maximum privacy loss on a ten point trace for a diagonal `GI' covariance matrix along with a series of other RBF matrices with different length scales. We observe $\mathbf{L}_{\text{priv}}^*$ for a subsequence $\calS$ of every other point, and see a 50\% increase in privacy loss even for $l=1$, which assigns a correlation of 0.6 to neighboring points. See \textbf{Figure \ref{gaussian figs}} for plots of this privacy loss. 

\newpage

\medskip

\bibliographystyle{plainnat}
\bibliography{refs.bib}

\newpage

\section{Appendix}
\subsection{Demonstration of Results}

\textbf{Corollary \ref{expected_gap}.}
\textit{
For two distributions, $P(\bZ | \bS_i, \theta), P(\bZ | \bS_j, \theta)$, with R\'enyi Divergence bounded by $\epsilon$, we also know that the expected gap between the prior and posterior log-odds ratios is bounded by $\epsilon$. 
}
	\begin{align*}
		\E_{\bZ \sim P(\bZ | \bS_j, \theta)} \bigg[ \log \frac{P(\bS_i | \bZ, \theta)}{P(\bS_j | \bZ, \theta)} - \log \frac{P(\bS_i | \theta)}{P(\bS_j | \theta)} \bigg] 
		&\leq D_{\lambda} \binom{P(\bZ | \bS_i, \theta)}{P(\bZ | \bS_j, \theta)}  \\
		&\leq \epsilon 
	\end{align*} 

\begin{proof}
	\begin{align*}
		\E_{\bZ \sim P(\bZ | \bS_j, \theta)} \bigg[ \bigg( \frac{P(\bZ | \bS_i, \theta)}{P(\bZ | \bS_j, \theta)} \bigg)^\lambda \bigg]
		&= \exp \bigg[ (\lambda - 1) D_\lambda \binom{P(\bZ | \bS_i, \theta)}{P(\bZ | \bS_j, \theta)} \bigg] \tag{a} \\
		\lambda \E_{\bZ \sim P(\bZ | \bS_j, \theta)} \big[ \log P(\bZ | \bS_i, \theta) - \log P(\bZ | \bS_j, \theta) \big] 
		&\leq (\lambda - 1) D_\lambda \binom{P(\bZ | \bS_i, \theta)}{P(\bZ | \bS_j, \theta)} \tag{b} \\
		 \E_{\bZ \sim P(\bZ | \bS_j, \theta)} \big[ \log P(\bZ | \bS_i, \theta) - \log P(\bZ | \bS_j, \theta) \big] 
		&\leq  D_\lambda \binom{P(\bZ | \bS_i, \theta)}{P(\bZ | \bS_j, \theta)} \tag{c} \\
		\E_{\bZ \sim P(\bZ | \bS_j, \theta)} \bigg[ \log \frac{P(\bS_i | \bZ, \theta)}{P(\bS_j | \bZ, \theta)} - \log \frac{P(\bS_i | \theta)}{P(\bS_j | \theta)} \bigg]
		&\leq  D_\lambda \binom{P(\bZ | \bS_i, \theta)}{P(\bZ | \bS_j, \theta)} \tag{d}
	\end{align*}
	Step (a) simply uses the definition of R\'enyi Divergence, which for two distributions $\mu, \nu$ defined on the same measurable space, and for order $\lambda > 1$ is given by 
	\begin{align*}
		D_\lambda \binom{\mu}{\nu} = \frac{1}{\lambda - 1} \log \E_{x \sim \nu} \bigg( \frac{\gamma(x)}{\nu(x)} \bigg)^\lambda
	\end{align*}
	Step (b) first takes the log of both sides, and then applies Jensen's Inequality. Step (c) uses the fact that $\frac{\lambda - 1}{\lambda} < 1 \ \forall \lambda > 1$. Step (d) simply uses Bayes Rule. 
\end{proof}

\textbf{Equation \ref{div terms}.}
\begin{align*} 
	\tag{\ref{div terms}}
	D_\lambda \binom{P(\bZ | \bS_i, \theta)}{P(\bZ | \bS_j, \theta)}  
	&= D_\lambda \binom{P(\bU_i + g_u)}{P(\bU_j + g_u)}  +
	\sum_{(z, x_i, x_j) \in (\calZ_s, \calS_i, \calS_j)} 
	D_\lambda \binom{P(z | x_i)}{P(z | x_j}  
\end{align*}
\begin{proof}
	\begin{align*}
		D_\lambda \binom{P(\bZ | \bS_i, \theta)}{P(\bZ | \bS_j, \theta)}  
		&= D_\lambda \binom{P(\bZ_s | \bS_i) P(\bZ_u | \bS_i, \theta)}{P(\bZ_s | \bS_j)P(\bZ_u | \bS_j, \theta)} \tag{a} \\
	\end{align*}
	Simplify $P(\bZ_u | \bS_i, \theta)$ according to the model defined in \textbf{Figure \ref{graphical models}}: 
	\begin{align*}
		P(\bZ_u | \bS_i, \theta) &= \int_{\R^{|\calU|}} P(\bZ_u, \bU = \bu | \bS_i, \theta) d\bu \\
		&= \int_{\R^{|\calU|}} P(\bU = \bu | \bS_i, \theta) P(\bZ_u | \bU = \bu) d\bu \\
		&= \int_{\R^{|\calU|}} P(\bU = \bu | \bS_i, \theta) P(g_u = \bZ_u - \bu | \zeta_u) d\bu \\
		&= \big[ P(\bU | \bS_i, \theta) * P(g_u | \zeta_u) \big] (\bZ_u) \\
		&= P(\bU_i + g_u)
	\end{align*}
	where $\bU_i \sim P(\bU | \bS_i)$ and $g_u \sim \zeta_u$. Returning to the divergence calculation at step (a)
	\begin{align*}
		D_\lambda \binom{P(\bZ_s | \bS_i) P(\bZ_u | \bS_i, \theta)}{P(\bZ_s | \bS_j)P(\bZ_u | \bS_j, \theta)} 
		&= D_\lambda \binom{P(\bZ_s | \bS_i) P(\bU_i + g_u)}{P(\bZ_s | \bS_j)P(\bU_j + g_u)} \tag{b} \\
		&= D_\lambda \binom{P(\bZ_s | \bS_i)}{P(\bZ_s | \bS_j)} + D_\lambda \binom{P(\bU_i + g_u)}{P(\bU_j + g_u)} \tag{c} \\
		&= \bigg[ \sum_{(z, x_i, x_j) \in (\bZ_s, \bS_i, \bS_j)}  D_\lambda \binom{P(z | x_i)}{P(z | x_j)} \bigg] + D_\lambda \binom{P(\bU_i + g_u)}{P(\bU_j + g_u)} \tag{d} \\
		&= D_\lambda \binom{P(\bU_i + g_u)}{P(\bU_j + g_u)} +  \sum_{(z, x_i, x_j) \in (\calZ_s, \calS_i, \calS_j)}  D_\lambda \binom{P(z | x_i)}{P(z | x_j)}
	\end{align*}
	Step (a) uses the conditional independence properties specified in \textbf{Figure \ref{graphical models}}. Step (b) substitutes the redefined distribution of $P(\bZ_u | \bS_i, \theta)$ into the divergence. Step (c) is possible because $\bZ_s \perp \bZ_u | \bS$ as specified in \textbf{Figure \ref{graphical models}}; this is a property of R\'enyi Divergence for product distributions. Step (d) separates the first term into $|\calS|$ separate terms since, each of the $z_i$s are independent conditioned on $\bS$, thus separating their R\'enyi Divergences into the sum of $|\calS|$ terms. 
\end{proof}

\textbf{Equation \ref{2terms}.}
\begin{align*}
	\mathbf{L}_{\text{priv}}
	=& \frac{\lambda}{2} \Big[ \Delta \bs^\intercal \Sigma_{\textit{eff}} \Delta \bs + 
	\frac{1}{\sigma_z^2} \| \Delta \bs \|_2^2  \Big] \tag{\ref{2terms}}
\end{align*}

\begin{proof}
	\begin{align*}
		\mathbf{L}_{\text{priv}}
		=&D_\lambda \binom{P(\bU_i + g_u)}{P(\bU_j + g_u)} +  \sum_{(z, x_i, x_j) \in (\bZ_s, \bS_i, \bS_j)}  D_\lambda \binom{P(z | x_i)}{P(z | x_j)} \tag{a} \\
		=&  D_\lambda \binom{\calN(\bmu_{u|s_i}, \Sigma_{u|s_i} + \sigma_z^2 \mathbb{I})}{\calN(\bmu_{u|s_j}, \Sigma_{u|s_j} + \sigma_z^2 \mathbb{I})} +
		\sum_{(x_i, x_j) \in (\calS_i, \calS_j)} D_\lambda \binom{\calN(x_i, \sigma_z^2)}{\calN(x_i, \sigma_z^2)}  \tag{b}\\
		=& \frac{\lambda}{2} (\bmu_{u|s_i} - \bmu_{u|s_j} )^{\intercal} (\Sigma_{u|s} + \sigma_z^2 \mathbb{I})^{-1} (\bmu_{u|s_i} - \bmu_{u|s_j}) +
		  \frac{\lambda }{2\sigma_z^2} \|\bs_i - \bs_j\|_2^2 \tag{c}
	\end{align*}
	Step (a) is simply a restatement of \textbf{Equation \ref{div terms}}. Step (b) substitutes in the distributions for a Gaussian conditional prior and Gaussian additive noise mechanism. Step (c) is due to the R\'enyi Divergence of two normal distributions with identical covariance, which is the Mahalanobis distance between the conditional means: 
	\begin{align*}
		D_\lambda \binom{\calN(\bmu_a, \Sigma)}{\calN(\bmu_b, \Sigma)} 
		= \frac{\lambda}{2}(\bmu_a - \bmu_b)^\intercal \Sigma^{-1} (\bmu_a - \bmu_b)
	\end{align*}
	In step (c), $\Sigma_{u|s} = \Sigma_{u|s_i} = \Sigma_{u|s_j}$ since Gaussian conditional covariance matrices only depend on \emph{which} points are being conditioned on, not their values. Substituting in the values of $\bmu_{u|s_i} = \Sigma_{us}\Sigma_{ss}^{-1}\bs_i$ and $\bmu_{u|s_j}= \Sigma_{us}\Sigma_{ss}^{-1}\bs_j$, which is defined by the Gaussian conditional mean, this expression simplifies nicely. Let $\Delta \bs = (\bs_i - \bs_j) $, then the above expression becomes 
	\begin{align*}
		=& \frac{\lambda}{2} \Big[ \Delta \bs^\intercal (\Sigma_{us}\Sigma_{ss}^{-1})^\intercal (\Sigma_{u|s} + \sigma_z^2 \mathbb{I})^{-1}
	     (\Sigma_{us}\Sigma_{ss}^{-1}) \Delta \bs + 
	     \frac{1}{\sigma_z^2} \| \Delta \bs \|_2^2 \Big] \tag{d}\\
	    =& \frac{\lambda}{2} \Big[ \Delta \bs^\intercal \Sigma_{\textit{eff}} \Delta \bs + 
	    \frac{1}{\sigma_z^2} \| \Delta \bs \|_2^2  \Big] \tag{e}
	\end{align*}
	where $\Sigma_{\textit{eff}} = (\Sigma_{us}\Sigma_{ss}^{-1})^\intercal (\Sigma_{u|s} + \sigma_z^2 \mathbb{I})^{-1} (\Sigma_{us}\Sigma_{ss}^{-1})$. One can see that increasing $\sigma_z^2$, reduces the eccentricity of $(\Sigma_{u|s} + \sigma_z^2 \mathbb{I})^{-1}$, and thus reduces the top eigenvalue and the worst case privacy loss. Meanwhile, having a longer length scale, $l$, in the kernel increases covariance, and increases the eccentricity of $(\Sigma_{u|s} + \sigma_z^2 \mathbb{I})^{-1}$. 
\end{proof}

\textbf{Equation \ref{Lpriv_star}.}
\begin{align}
	\mathbf{L}_{\text{priv}}^* &= \frac{\lambda}{2} \bigg( \frac{1 + \sigma_z^2 \alpha^*}{\sigma_z^2} \bigg) |\calS|  r^2 \tag{\ref{Lpriv_star}}
\end{align}
\begin{proof}
	\begin{align*}
		\mathbf{L}_{\text{priv}} 
		&= \frac{\lambda}{2} \Big[ \Delta \bs^\intercal \Sigma_{\textit{eff}} \Delta \bs + 
	    \frac{1}{\sigma_z^2} \| \Delta \bs \|_2^2  \Big] \tag{a} \\
	    \mathbf{L}_{\text{priv}}^* 
	    &= \max_{\Delta \bs \in \Delta \calS} 
	    \frac{\lambda}{2} \Big[ \Delta \bs^\intercal \Sigma_{\textit{eff}} \Delta \bs + 
	    \frac{1}{\sigma_z^2} \| \Delta \bs \|_2^2  \Big] \tag{b} \\
	    &= \frac{\lambda}{2} \Big[ \Delta \bs^{*\intercal} \Sigma_{\textit{eff}} \Delta \bs^* + 
	    \frac{1}{\sigma_z^2} \| \Delta \bs^* \|_2^2  \Big] \tag{c} \\
	    &= \frac{\lambda}{2} \Big[ \alpha^* \| \Delta \bs^* \|_2^2 + 
	    \frac{1}{\sigma_z^2} \| \Delta \bs^* \|_2^2  \Big] \tag{d} \\
	    &= \frac{\lambda}{2} \bigg( \frac{1 + \sigma_z^2 \alpha^*}{\sigma_z^2} \bigg)\| \Delta \bs^* \|_2^2 \tag{e} \\
	    &= \frac{\lambda}{2} \bigg( \frac{1 + \sigma_z^2 \alpha^*}{\sigma_z^2} \bigg)|\calS| r^2 \tag{f}
	\end{align*}
	Step (a) is a restatement of \textbf{Equation \ref{2terms}}. Step (b) is a maximization of all discriminative pairs described in Definition \ref{CIP def}. Here, the set of all pairs is given by $\Delta \calS = \{\Delta \bs = \bs_i - \bs_j : \|\Delta \bs\|_2^2 \leq |\calS| r^2\}$. The choice of $|\calS| r^2$ as the upper bound on magnitude, is because this includes a radius of $r$ around each $x_i \in \calS$. Step (c) substitutes in the optimal solution $\Delta \bs^*$. Step (d) identifies the optimal solution as the maximizer of the Mahalanobis distance of the first term. Here, $\alpha^*$ is the top eigenvalue of $\Sigma_{\textit{eff}}$. The maximizer $\Delta \bs^*$ will also have maximum magnitude, $\|\Delta \bs^*\| = |\calS| r^2$, in the direction of the top eigenvector, which is incorporated into step (f). 
\end{proof}

\end{document}